\documentclass{article}

\usepackage[final]{neurips_2020_ml4ps}
\usepackage[utf8]{inputenc} 
\usepackage[T1]{fontenc}    
\usepackage{hyperref}       
\usepackage{url}            
\usepackage{booktabs}       
\usepackage{amsfonts}       
\usepackage{nicefrac}       
\usepackage{microtype}      
\usepackage{graphicx}
\usepackage{multirow}

\title{Simultaneously forecasting global geomagnetic activity using recurrent networks}

\author{%
  Charles Topliff \\
  Georgia Institute of Technology\\
  Atlanta, GA 30332 \\
  \texttt{ctopliff0@gatech.edu} \\
   \And
   Morris Cohen \\
   Georgia Institute of Technology \\
   Atlanta, GA 30332 \\
   \texttt{mcohen@gatech.edu} \\
   \AND
   William Bristow \\
   Pennsylvania State University \\
   State College, PA 16801 \\
   \texttt{wab5217@psu.edu} \\

}

\begin{document}

\maketitle

\begin{abstract}

   Many systems used by society are extremely vulnerable to space weather events such as solar flares and geomagnetic storms which could potentially cause catastrophic damage. In recent years, many works have emerged to provide early warning to such systems by forecasting these events through some proxy, but these approaches have largely focused on a specific phenomenon. We present a sequence-to-sequence learning approach to the problem of forecasting global space weather conditions at an hourly resolution. This approach improves upon other work in this field by simultaneously forecasting several key proxies for geomagnetic activity up to 6 hours in advance. We demonstrate an improvement over the best currently known predictor of geomagnetic storms, and an improvement over a persistence baseline several hours in advance.

\end{abstract}

\vspace{-.1in}
\section{Introduction}

 Many ground systems are influenced or disrupted by extreme space weather, such as power infrastructure, satellite communication, television and radio links, as well as GPS navigation systems \citep{allen1989effects, kan1979energy, ding2007large, board2009severe}. Forecasts of space weather events would provide early warning to these systems, allowing for precautions to be taken to mitigate the effects of otherwise damaging activity. One area of particular interest is the modeling of variations in Earth's magnetic field, or geomagnetic activity. There exist many different classes of geomagnetic activity, such as magnetic storms, magnetospheric substorms, quiet day variation, etc \citep{chapman1962geomagnetism}. The physical processes behind many of these phenomena begin in the interior of the sun, where pressure builds to a release of a steady stream of radiation and energetic particles known as the solar wind. Geomagnetic storms at Earth are often the result of periods of drastically intensified solar wind known as coronal mass ejections. In recent years, the problem of forecasting such storms has gained much attention. Geomagnetic indices were developed to measure the general intensity of space weather from the ground \cite{kivelson1995introduction}. Recent works focus on modeling a particular class of geomagnetic activity by studying the relationship between solar wind measurements obtained from space weather datasets or derived from tertiary datasets, and a proxy for magnetic activity such as magnetic indices which summarize specific geomagnetic phenomena.
 
  In this work, we seek to predict multiple magnetic indices such as the auroral electrojet indices (AE, AU, AL), the disturbance time index (Dst), the Solar Radio Flux at 10.7cm (F10.7) index, and the Planetary K (Kp) index several hours in advance. Where the majority of works have focused on predicting a single magnetic index, we demonstrate an approach to simultaneously forecast several indices multiple hours in advance. In our work, we frame the problem as a multivariate sequence-to-sequence learning task \citep{sutskever2014sequence} where the objective is to predict the four magnetic indices using solar wind measurements. We utilize long short-term memory networks which have been applied to myriad applications involving sequential data and are known for addressing vanishing gradient issues encountered with vanilla recurrent neural networks \citep{hochreiter1997long}. 

\vspace{-.1in}
\section{Related Work}

\vspace{-.1in}
\subsection{Magnetic Index Forecasting}

Many works focus on modeling a particular class of geomagnetic activity by examining a single magnetic index. One example area of research is that of studying solar wind coupling where the objective is to compute some coupling function and solve a least-squares regression problem to predict a particular magnetic index using the time-lagged coupling function. \citet{mcpherron2015optimum} introduced a coupling function which partially linearizes the relationship between a set of commonly used solar wind variables and solved a least-squares regression problem to predict the AL index one hour in advance. They reported an average coefficient of determination ($R^2$) of $.67$ over two-month periods spanning 1963-2014 which improves upon other solar wind coupling functions \citep{newell2007nearly, rostoker1972response, clauer1986relationship}.

Other works which utilize neural networks have emerged in recent years. \citet{revallo2014neural} investigated forecasting the Dst index one hour in advance using neural networks, and \citet{boberg2000real} sought to predict the Kp index in real time using solar wind measurements. In \citet{maimaiti2019deep}, the authors derived a substorm classification dataset by applying the substorm identification criteria \citep{newell2011substorm} to the SuperMAG SML index \citep{gjerloev2012supermag}. The forecasting approach involves training variations of (previously) state-of-the-art convolutional networks, such as ResNet \citep{he2016identity} to predict the onset of a magnetospheric substorm within the next hour.

\vspace{-.1in}
\subsection{Solar Wind Forecasting}
Another area of research is the prediction of Solar Wind measurements at L1. NASA's solar dynamics observatory curates a database of ultraviolet images of the solar corona \citep{galvez2019machine} which has been used by many works to forecast measurements at L1. Works typically model complex coupling processes which occur between the Sun and the Earth's magnetic field in order to predict solar wind speed (V) or interplanetary magnetic field strength (B). \citet{upendran2020solar} used a CNN-LSTM architecture leveraging a GoogLeNet \citep{szegedy2015going} as a feature extractor to forecast daily averages of the solar wind multiple days in advance. \citet{chandorkar2019dynamic} proposed a new regression framework to model the varying time lag implicit to the coupling processes between the solar magnetic field and solar wind propagation around the heliosphere.

\vspace{-.1in}
\section{Data and Preprocessing}

\vspace{-.1in}
\subsection{Solar wind and IMF Measurements}
 We use measurements of the solar wind speed $V$, proton density $n$, and the components of the interplanetary magnetic field (IMF) $B_x, B_y, B_z$.  These measurements are taken by the ACE and WIND satellites located at L1. We also utilize historical values of the magnetic indices AE, AL, AU, Dst, F10.7, and Kp. This data is curated by NASA's Space Physics Data Facility and is available through the OMNIWeb database\footnote{OMNIWeb data available at https://omniweb.gsfc.nasa.gov/}. As additional features, we include calendar year, day of year, and hour of day signals. These signals are converted to sinusoids with periods 11 years (roughly equivalent to the average length of a solar cycle), 365 days, and 24 hours in order to account for cyclical variations in the data. All data available on the OMNIWeb database has been preprocessed and resampled to several different time resolutions. For our experiments, we utilize hourly resolution data. 

\vspace{-.1in}
\subsection{SuperDARN Measurements}

We also utilize measurements derived from the SuperDARN network of radars \citep{greenwald1995darn}. Specifically, we used the spherical harmonic fit technique developed by \citet{ruohoniemi1989mapping} to produce a database of convection patterns covering the years from 2013 to 2017 (inclusive) on a five-minute cadence. These measurements were subsequently averaged and resampled to a 1 hour resolution. From these patterns we extracted two parameters, the Cross-Polar-Cap Potential (CCP), which parameterizes the degree to which solar wind energy is coupled into the Earth’s magnetosphere, (hence is one of the best parameters available for describing the level of global-scale magnetospheric activity), and the Polar Cap Radius (PCR), which is the average colatitude of the Convection Reversal Boundary. These measurements provide the best proxy for magnetic conditions at the poles aside from the auroral electrojet indices themselves. For a more detailed discussion, the reader is referred to \citep{bristow2007superposed}.

\vspace{-.1in}
\section{Model Training}

\vspace{-.1in}
\subsection{Dataset Assembly}
We treat this forecasting problem as a multivariate sequence-to-sequence learning problem. At some time $t$, the inputs to our model comprise the samples of the time series of all input features in the interval $[t - T_h, t]$ where $T_h \geq 0$ is the number of samples in the past considered. The target is the samples of the time series of the four magnetic indices in the interval $(t, t + T_p]$ where $T_p \geq 1$ is the number of samples in the future predicted. Figure 1 illustrates the structure of input/output pairs on a subset of features. To assemble our datasets, we simply slide this structure through the entire time series one sample at a time. For all of our experiments, we discard all datapoints with missing values. To avoid violating causality, we split the data into training, validation, and testing sets sequentially with sizes being roughly 60, 10, and 30 percent, respectively. We center and scale the input features and training labels by the mean and standard deviation of each indiviual feature in the training set. We perform no additional preprocessing or filtering by the magnitude of the magnetic indices.

\begin{figure}[t]
  \centering
  \includegraphics[width=0.9 \textwidth]{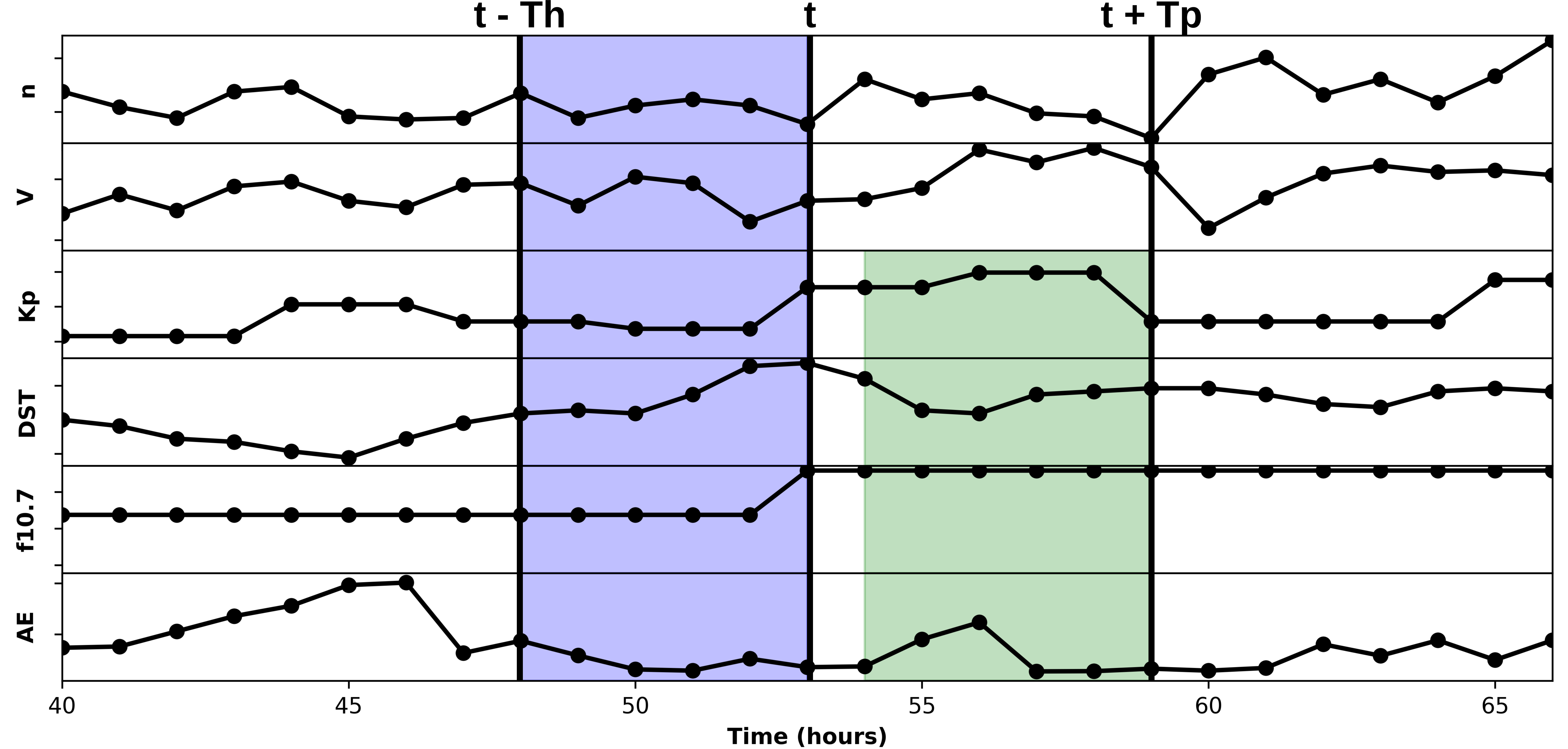}
  \caption{Structure of a single datum in our dataset. The blue shaded region between $t-T_h$ and $t$ comprises all samples in the input time series and the green shaded region between $t$ and $t + T_p$ comprises all samples in the target time series. In this example, $T_p = 6, T_h = 5$.}
\end{figure}

\vspace{-.1in}
\subsection{Model Training and Hyperparameter Tuning}
In this work, we consider a simple neural network consisting of a LSTM which embeds the input time series into a low dimensional space followed by a linear layer which performs a linear transformation on the final output hidden state of the LSTM to the dimensionality of the output time series. We use Mean Squared Error as our loss function, and optimize the weights of the network using the Adam optimizer \citep{kingma2014adam} with weight decay. We conduct all experiments using the PyTorch deep learning framework \citep{NEURIPS2019_9015}. We treat the optimizer learning rate, weight decay coefficient, batch size, hidden dimension, and the number of layers of the LSTM as training hyperparameters. For each experiment, we conduct a random search \citep{bergstra2012random} over 100 total models and train the models for 1350 epochs using the median early stopping rule. We select the best model ranked by the performance on the validation set for all experiments.

\vspace{-.1in}
\section{Results}
\textit{20 Year OMNIWeb Data} In this experiment, we use 20 contiguous years of OMNIWeb Data (without including SuperDARN data) ranging from the years 2000-2019. We use a time history and a lead time of $T_h = T_p = 6$ hours. Our training, validation, and testing datasets correspond roughly to the years 2000-2011, 2012-2013, and 2014-2019, respectively. Our baseline for these exeriments is a persistence forecast in which the prediction for the magnetic indices for all $t$ is equivalent to the most recent available observation in the historical indices. In this experiment, we report performance in terms of pearson corellation coefficient $\rho$ computed over the testing set. Table 1 illustrates the performance drop off over time for our network (mNet) and a persistence forecast (pers). The purpose of this experiment is to compare to a baseline persistent forecast, but it should be noted that we also compare to \citep{mcpherron2015optimum} by computing the coefficient of determination (or explained variance) for a 1 hour AL forecast using predictions from the same network. We find that we achieve an $R^2$ of $.744$ over our test set ranging from 2014-2019. For this experiment, the best performing model uses a one layer LSTM, a batch size of 502, hidden dimension of 58, an initial learning rate of .00368, and a weight decay parameter of 9.7E-8.

\begin{table}[h]
\small
\caption{Pearson correlation coefficent for mNet and baseline (pers) predictions with test labels.}
\label{sample-table}
\begin{tabular}{ccccccccccccc}
\hline
 & \multicolumn{2}{c}{\textbf{AE}} & \multicolumn{2}{c}{\textbf{AU}} & \multicolumn{2}{c}{\textbf{AL}} & \multicolumn{2}{c}{\textbf{Dst}} & \multicolumn{2}{c}{\textbf{f10.7}} & \multicolumn{2}{c}{\textbf{Kp}} \\ \hline
\textbf{Hrs} & \textbf{mNet} & \textbf{pers} & \textbf{mNet} & \textbf{pers} & \textbf{mNet} & \textbf{pers} & \textbf{mNet} & \textbf{pers} & \textbf{mNet} & \textbf{pers} & \textbf{mNet} & \textbf{pers} \\
\hline
1 & .901 & .478 & .885 & .491 & .874 & .422 & .976 & .790 & .995 & .993 & .937 & .658 \\
2 & .782 & .452 & .789 & .462 & .743 & .397 & .949 & .759 & .994 & .991 & .882 & .632 \\
3 & .696 & .429 & .715 & .437 & .656 & .375 & .918 & .730 & .993 & .990 & .825 & .605 \\
4 & .650 & .406 & .665 & .414 & .616 & .352 & .892 & .703 & .992 & .989 & .790 & .578 \\
5 & .622 & .382 & .627 & .394 & .594 & .328 & .870 & .678 & .991 & .988 & .757 & .557 \\
6 & .598 & .364 & .598 & .379 & .574 & .309 & .850 & .655 & .990 & .987 & .726 & .535 \\
\hline
\end{tabular}
\end{table}

\textit{SuperDARN measurements} We compare the performance of our model on three different datasets to examine the added predictive capability of SuperDARN measurements. We start with the original OMNIWeb dataset as described in the previous experiment (base), create an additional dataset by replacing the historical values of the magnetic indices with the SuperDARN measurements (sdrn), and finally create a baseline dataset where we only use solar wind measurements (sw). The hyperparameters for the selected models are given in Table 2. For a fair comparison, we discard all datum corresponding to hours which are not represented in all three datasets. Similar to the first experiment, we use a time history and a lead time of $T_h = T_p = 6$ hours. Figure 2 illustrates the performance of the three models in terms of the pearson correlation coefficient of the predictions with test labels. 

\begin{figure}[b]
  \centering
  \includegraphics[scale=.28]{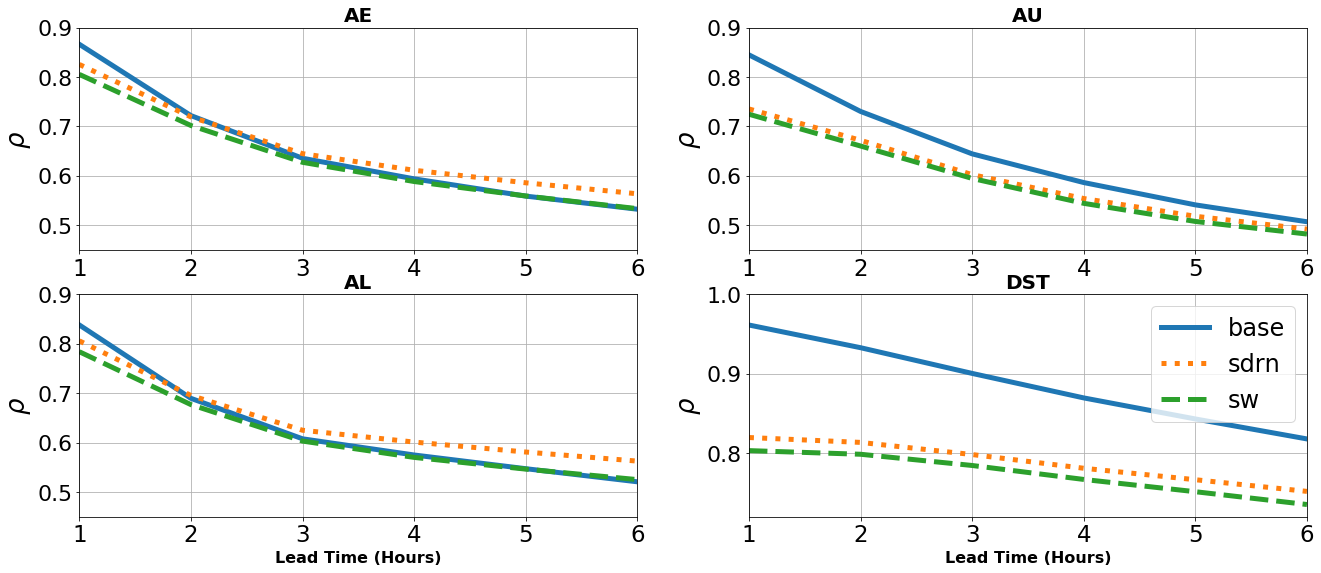}
  \caption{Pearson correlation for the predictions of the base model (blue/solid), sdrn model (orange/dotted), and sw model(red/dashed) with test labels.}
\end{figure}

We find that the SuperDARN measurements alone do not provide a substantial improvement over the base solar wind model, and thus we cannot conclude on their efficacy as a sufficient proxy for the magnetic indices. We also remark that the inconclusive results may be due to the limited data availability of SuperDARN measurements. To this end, we hope to increase the amount of SuperDARN measurements in future work.

\begin{table}[t]
\small
\centering
\caption{Hyperparameters for the models in the \textit{SuperDARN measurements} experiment.}
\begin{tabular}{cccccc}
\hline
\textbf{Model} & \textbf{LSTM Layers} & \textbf{Weight Decay} & \textbf{Hidden Dimension} & \textbf{Batch Size} & \textbf{Learning Rate} \\
\hline
\textbf{base} & 1 & 4.21E-8 & 59 & 1731 & 4.73E-2 \\
\textbf{sw} & 1 & 2.31E-7 & 44 & 1032 & 4.79E-2 \\
\textbf{sdrn} & 1 & 4.21E-2 & 48 & 1060 & 2.87E-6 \\ \hline
\end{tabular}
\end{table}

\vspace{-.1in}
\section{Conclusion}
\vspace{-.1in}

We have presented a sequence-to-sequence learning approach to forecasting several geomagnetic indices simultaneously. These results demonstrate the efficacy of such an approach to forecasting these magnetic indices, an improvement over the current state-of-the-art for magnetospheric substorm prediction as well as an ability to outperform a persistence forecast over a longer time horizon. We find that the addition of the currently available SuperDARN data does not improve the predictive power of our models substantially. We seek to include additional results after more SuperDARN data has been processed in the future. Much additional future work is planned. For example, a larger model class may result in improved forecasts. Many recent works have emerged for similar time series forecasting problems including the use of attention, one-dimensional convolutional networks, and other recurrent neural networks. We seek to investigate these by performing a model architecture search \citep{elsken2018neural} or by using pretrained models similar to the approach used in \citep{maimaiti2019deep}. Additionally, prior works have used the ultraviolet coronal image dataset curated by NASA to forecast the solar wind and magnetic conditions at L1 orbit point at a coarse granularity. One potential future direction is to utilize a similar approach to forecast the magnetic indices directly and model the Earth-sun system in an end-to-end fashion. Such an approach may benefit from a pretrained convolutional network. Lastly, we seek to investigate the application of physics-informed neural networks to this problem by exploiting domain knowledge as a regularizer, such as the relationship between the AE, AL, and AU indices.

\vspace{-.1in}
\section*{Broader Impact}
\vspace{-.1in}
Space weather is a broad term used to describe many potentially catastrophic phenomena for which we currently have limited forecasting capability. We know such events will happen, we know the consequences can be disastrous, but we do not have the sophistication to predict when or how a particular event will evolve. One example is a famous 1859 `Carrington Event' that occurred during which northern lights were visible from Central America. Communication infrastructure was not nearly as advanced as it is today, but early telegraph stations caught on fire from induced electrical activity. The same thing could happen today to our electric power grid. The National Academies of Sciences estimates that an event of similar magnitude occurring today could cause several trillion dollars, knocking out power for many countries for many months. Currently, alerts to power utility companies about potential space weather disruptions are based exclusively on these geomagnetic indices, thus improving forecasts of these indices would immediately provide a tangible benefit to society. Space weather forecasting remains in its infancy, in part due to the cacophony of physical processes that cascade from the Sun to the Earth when a solar storm occurs. For decades, physicists have investigated this problem and have made substantial advances in modeling individual components of these complex systems, but have struggled to model the system with analytical techniques in an end-to-end fashion. As such, the broad impacts of effectively applying machine learning to the space weather forecasting problem could be huge, as ML techniques have a unique ability to describe systems that cannot be modeled analytically.
    
\vspace{-.1in}
\begin{ack}
This work benefited directly from conversations with Jacob Bortnik and Ryan McGranaghan. We would like to thank them for their contributions.

This work was supported by the Defense Advanced Research Projects Agency (DARPA) under cooperative agreement D19AC00009 from the Department of the Interior to Georgia Tech. The first author is supported by the NDSEG fellowship.
\end{ack}

\bibliographystyle{unsrtnat}
\bibliography{ref}

\begin{thebibliography}{29}
\providecommand{\natexlab}[1]{#1}
\providecommand{\url}[1]{\texttt{#1}}
\expandafter\ifx\csname urlstyle\endcsname\relax
  \providecommand{\doi}[1]{doi: #1}\else
  \providecommand{\doi}{doi: \begingroup \urlstyle{rm}\Url}\fi

\bibitem[Allen et~al.(1989)Allen, Sauer, Frank, and Reiff]{allen1989effects}
Joe Allen, Herb Sauer, Lou Frank, and Patricia Reiff.
\newblock Effects of the march 1989 solar activity.
\newblock \emph{Eos, Transactions American Geophysical Union}, 70\penalty0
  (46):\penalty0 1479--1488, 1989.

\bibitem[Kan and Lee(1979)]{kan1979energy}
JR~Kan and LC~Lee.
\newblock Energy coupling function and solar wind-magnetosphere dynamo.
\newblock \emph{Geophysical Research Letters}, 6\penalty0 (7):\penalty0
  577--580, 1979.

\bibitem[Ding et~al.(2007)Ding, Wan, Ning, and Wang]{ding2007large}
F~Ding, W~Wan, B~Ning, and M~Wang.
\newblock Large-scale traveling ionospheric disturbances observed by gps total
  electron content during the magnetic storm of 29--30 october 2003.
\newblock \emph{Journal of Geophysical Research: Space Physics}, 112\penalty0
  (A6), 2007.

\bibitem[Board et~al.(2009)Board, Council, et~al.]{board2009severe}
Space~Studies Board, National~Research Council, et~al.
\newblock \emph{Severe space weather events: Understanding societal and
  economic impacts: A workshop report}.
\newblock National Academies Press, 2009.

\bibitem[Chapman and Bartels(1962)]{chapman1962geomagnetism}
S~Chapman and J~Bartels.
\newblock Geomagnetism, 2nd edn., vol. 1.
\newblock \emph{Clarendon, Oxford}, 1962.

\bibitem[Kivelson et~al.(1995)Kivelson, Kivelson, and
  Russell]{kivelson1995introduction}
Margaret~G Kivelson, Margaret~Galland Kivelson, and Christopher~T Russell.
\newblock \emph{Introduction to space physics}.
\newblock Cambridge university press, 1995.

\bibitem[Sutskever et~al.(2014)Sutskever, Vinyals, and
  Le]{sutskever2014sequence}
Ilya Sutskever, Oriol Vinyals, and Quoc~V Le.
\newblock Sequence to sequence learning with neural networks.
\newblock In \emph{Advances in neural information processing systems}, pages
  3104--3112, 2014.

\bibitem[Hochreiter and Schmidhuber(1997)]{hochreiter1997long}
Sepp Hochreiter and J{\"u}rgen Schmidhuber.
\newblock Long short-term memory.
\newblock \emph{Neural computation}, 9\penalty0 (8):\penalty0 1735--1780, 1997.

\bibitem[McPherron et~al.(2015)McPherron, Hsu, and Chu]{mcpherron2015optimum}
Robert~L McPherron, Tung-Shin Hsu, and Xiangning Chu.
\newblock An optimum solar wind coupling function for the al index.
\newblock \emph{Journal of Geophysical Research: Space Physics}, 120\penalty0
  (4):\penalty0 2494--2515, 2015.

\bibitem[Newell et~al.(2007)Newell, Sotirelis, Liou, Meng, and
  Rich]{newell2007nearly}
PT~Newell, T~Sotirelis, K~Liou, C-I Meng, and FJ~Rich.
\newblock A nearly universal solar wind-magnetosphere coupling function
  inferred from 10 magnetospheric state variables.
\newblock \emph{Journal of Geophysical Research: Space Physics}, 112\penalty0
  (A1), 2007.

\bibitem[Rostoker et~al.(1972)Rostoker, Lam, and Hume]{rostoker1972response}
Gordon Rostoker, Hing-Lan Lam, and William~D Hume.
\newblock Response time of the magnetosphere to the interplanetary electric
  field.
\newblock \emph{Canadian Journal of Physics}, 50\penalty0 (6):\penalty0
  544--547, 1972.

\bibitem[Clauer and Banks(1986)]{clauer1986relationship}
C~Robert Clauer and Peter~M Banks.
\newblock Relationship of the interplanetary electric field to the
  high-latitude ionospheric electric field and currents: Observations and model
  simulation.
\newblock \emph{Journal of Geophysical Research: Space Physics}, 91\penalty0
  (A6):\penalty0 6959--6971, 1986.

\bibitem[Revallo et~al.(2014)Revallo, Valach, Hejda, and
  Bochn{\'\i}{\v{c}}ek]{revallo2014neural}
Milo{\v{s}} Revallo, Fridrich Valach, Pavel Hejda, and Josef
  Bochn{\'\i}{\v{c}}ek.
\newblock A neural network dst index model driven by input time histories of
  the solar wind--magnetosphere interaction.
\newblock \emph{Journal of Atmospheric and Solar-Terrestrial Physics},
  110:\penalty0 9--14, 2014.

\bibitem[Boberg et~al.(2000)Boberg, Wintoft, and Lundstedt]{boberg2000real}
Fredrik Boberg, Peter Wintoft, and Henrik Lundstedt.
\newblock Real time kp predictions from solar wind data using neural networks.
\newblock \emph{Physics and Chemistry of the Earth, Part C: Solar, Terrestrial
  \& Planetary Science}, 25\penalty0 (4):\penalty0 275--280, 2000.

\bibitem[Maimaiti et~al.(2019)Maimaiti, Kunduri, Ruohoniemi, Baker, and
  House]{maimaiti2019deep}
M~Maimaiti, B~Kunduri, JM~Ruohoniemi, JBH Baker, and Leanna~L House.
\newblock A deep learning-based approach to forecast the onset of magnetic
  substorms.
\newblock \emph{Space Weather}, 17\penalty0 (11):\penalty0 1534--1552, 2019.

\bibitem[Newell and Gjerloev(2011)]{newell2011substorm}
PT~Newell and JW~Gjerloev.
\newblock Substorm and magnetosphere characteristic scales inferred from the
  supermag auroral electrojet indices.
\newblock \emph{Journal of Geophysical Research: Space Physics}, 116\penalty0
  (A12), 2011.

\bibitem[Gjerloev(2012)]{gjerloev2012supermag}
JW~Gjerloev.
\newblock The supermag data processing technique.
\newblock \emph{Journal of Geophysical Research: Space Physics}, 117\penalty0
  (A9), 2012.

\bibitem[He et~al.(2016)He, Zhang, Ren, and Sun]{he2016identity}
Kaiming He, Xiangyu Zhang, Shaoqing Ren, and Jian Sun.
\newblock Identity mappings in deep residual networks.
\newblock In \emph{European conference on computer vision}, pages 630--645.
  Springer, 2016.

\bibitem[Galvez et~al.(2019)Galvez, Fouhey, Jin, Szenicer, Mu{\~n}oz-Jaramillo,
  Cheung, Wright, Bobra, Liu, Mason, et~al.]{galvez2019machine}
Richard Galvez, David~F Fouhey, Meng Jin, Alexandre Szenicer, Andr{\'e}s
  Mu{\~n}oz-Jaramillo, Mark~CM Cheung, Paul~J Wright, Monica~G Bobra, Yang Liu,
  James Mason, et~al.
\newblock A machine-learning data set prepared from the nasa solar dynamics
  observatory mission.
\newblock \emph{The Astrophysical Journal Supplement Series}, 242\penalty0
  (1):\penalty0 7, 2019.

\bibitem[Upendran et~al.(2020)Upendran, Cheung, Hanasoge, and
  Krishnamurthi]{upendran2020solar}
Vishal Upendran, Mark~CM Cheung, Shravan Hanasoge, and Ganapathy Krishnamurthi.
\newblock Solar wind prediction using deep learning.
\newblock \emph{Space Weather}, page e2020SW002478, 2020.

\bibitem[Szegedy et~al.(2015)Szegedy, Liu, Jia, Sermanet, Reed, Anguelov,
  Erhan, Vanhoucke, and Rabinovich]{szegedy2015going}
Christian Szegedy, Wei Liu, Yangqing Jia, Pierre Sermanet, Scott Reed, Dragomir
  Anguelov, Dumitru Erhan, Vincent Vanhoucke, and Andrew Rabinovich.
\newblock Going deeper with convolutions.
\newblock In \emph{Proceedings of the IEEE conference on computer vision and
  pattern recognition}, pages 1--9, 2015.

\bibitem[Chandorkar et~al.(2019)Chandorkar, Furtlehner, Poduval, Camporeale,
  and Sebag]{chandorkar2019dynamic}
Mandar Chandorkar, Cyril Furtlehner, Bala Poduval, Enrico Camporeale, and
  Michele Sebag.
\newblock Dynamic time lag regression: Predicting what \& when.
\newblock In \emph{International Conference on Learning Representations}, 2019.

\bibitem[Greenwald et~al.(1995)Greenwald, Baker, Dudeney, Pinnock, Jones,
  Thomas, Villain, Cerisier, Senior, Hanuise, et~al.]{greenwald1995darn}
RA~Greenwald, KB~Baker, JR~Dudeney, M~Pinnock, TB~Jones, EC~Thomas, J-P
  Villain, J-C Cerisier, C~Senior, C~Hanuise, et~al.
\newblock Darn/superdarn.
\newblock \emph{Space Science Reviews}, 71\penalty0 (1-4):\penalty0 761--796,
  1995.

\bibitem[Ruohoniemi et~al.(1989)Ruohoniemi, Greenwald, Baker, Villain, Hanuise,
  and Kelly]{ruohoniemi1989mapping}
JM~Ruohoniemi, RA~Greenwald, KB~Baker, J-P Villain, C~Hanuise, and J~Kelly.
\newblock Mapping high-latitude plasma convection with coherent hf radars.
\newblock \emph{Journal of Geophysical Research: Space Physics}, 94\penalty0
  (A10):\penalty0 13463--13477, 1989.

\bibitem[Bristow and Jensen(2007)]{bristow2007superposed}
WA~Bristow and Poul Jensen.
\newblock A superposed epoch study of superdarn convection observations during
  substorms.
\newblock \emph{Journal of Geophysical Research: Space Physics}, 112\penalty0
  (A6), 2007.

\bibitem[Kingma and Ba(2014)]{kingma2014adam}
Diederik~P Kingma and Jimmy Ba.
\newblock Adam: A method for stochastic optimization.
\newblock \emph{arXiv preprint arXiv:1412.6980}, 2014.

\bibitem[Paszke et~al.(2019)Paszke, Gross, Massa, Lerer, Bradbury, Chanan,
  Killeen, Lin, Gimelshein, Antiga, Desmaison, Kopf, Yang, DeVito, Raison,
  Tejani, Chilamkurthy, Steiner, Fang, Bai, and Chintala]{NEURIPS2019_9015}
Adam Paszke, Sam Gross, Francisco Massa, Adam Lerer, James Bradbury, Gregory
  Chanan, Trevor Killeen, Zeming Lin, Natalia Gimelshein, Luca Antiga, Alban
  Desmaison, Andreas Kopf, Edward Yang, Zachary DeVito, Martin Raison, Alykhan
  Tejani, Sasank Chilamkurthy, Benoit Steiner, Lu~Fang, Junjie Bai, and Soumith
  Chintala.
\newblock Pytorch: An imperative style, high-performance deep learning library.
\newblock In H.~Wallach, H.~Larochelle, A.~Beygelzimer, F.~d'Alch\'{e} Buc,
  E.~Fox, and R.~Garnett, editors, \emph{Advances in Neural Information
  Processing Systems 32}, pages 8024--8035. Curran Associates, Inc., 2019.
\newblock URL
  \url{http://papers.neurips.cc/paper/9015-pytorch-an-imperative-style-high-performance-deep-learning-library.pdf}.

\bibitem[Bergstra and Bengio(2012)]{bergstra2012random}
James Bergstra and Yoshua Bengio.
\newblock Random search for hyper-parameter optimization.
\newblock \emph{The Journal of Machine Learning Research}, 13\penalty0
  (1):\penalty0 281--305, 2012.

\bibitem[Elsken et~al.(2018)Elsken, Metzen, and Hutter]{elsken2018neural}
Thomas Elsken, Jan~Hendrik Metzen, and Frank Hutter.
\newblock Neural architecture search: A survey.
\newblock \emph{arXiv preprint arXiv:1808.05377}, 2018.

\end{thebibliography}

\end{document}